\documentclass[english]{IEEEtran}
\usepackage[T1]{fontenc}
\usepackage[latin9]{inputenc}
\usepackage{units}
\usepackage{amsmath}

\makeatletter

\usepackage{times}
\usepackage{graphicx}
\usepackage[export]{adjustbox}
\usepackage[english]{babel} 

\makeatletter
\newcommand\subparagraph{%
  \@startsection{subparagraph}{5}
  {\parindent}
  {3.25ex \@plus 1ex \@minus .2ex}
  {-1em}
  {\normalfont\normalsize\bfseries}}
\makeatother
\usepackage{titlesec}
\let\subparagraph\relax %

\usepackage{setspace}
\usepackage{multirow}
\usepackage{tabularx}
\usepackage{booktabs}
\usepackage{comment}
\usepackage{color, colortbl}
\usepackage{xcolor}
\usepackage{algorithm}
\usepackage{algpseudocode} 

\usepackage{caption}
\usepackage{subcaption}

\makeatletter
\newcommand*{\mycaptionfont}{\@setfontsize\mycaptionfont{8pt}{.85em}}
\makeatother

\DeclareCaptionFont{mycaptionfont}{\mycaptionfont}

\definecolor{figcapblue}{rgb}{0,0,0.4}
\algrenewcommand{\alglinenumber}[1]{\color{gray!80!gray}\tiny#1:}

\algtext*{EndIf}

\newcommand\NoDo{\renewcommand\algorithmicdo{}}

\newcommand\NoThen{\renewcommand\algorithmicthen{}}

\DeclareMathSizes{10}{8}{7}{5}
\DeclareMathSizes{11}{8}{7}{5}
\DeclareMathSizes{12}{8}{7}{5}

\definecolor{brightgreen}{rgb}{0.4, 1.0, 0.0}
\definecolor{brightlavender}{rgb}{0.75, 0.58, 0.89}
\definecolor{cadmiumred}{rgb}{0.89, 0.0, 0.13}
\definecolor{candyapplered}{rgb}{1.0, 0.03, 0.0}
\definecolor{amethyst}{rgb}{0.6, 0.4, 0.8}
\definecolor{aureolin}{rgb}{0.99, 0.93, 0.0}

\titlespacing\section{0pt}{5pt plus 3pt minus 3pt}{5pt plus 3pt minus 3pt}
\titlespacing\subsection{0pt}{2pt plus 2pt minus 0pt}{0pt plus 2pt minus 0pt}
\titlespacing\subsubsection{0pt}{2pt plus 2pt minus 0pt}{0pt plus 2pt minus 0pt}

\expandafter\def\expandafter\normalsize\expandafter{%
    \normalsize
    \setlength\abovedisplayskip{-9pt plus 2 pt minus 2pt}
    \setlength\belowdisplayskip{4pt plus 0 pt minus 0pt}
    \setlength\abovedisplayshortskip{-9pt plus 4 pt minus 2pt}
    \setlength\belowdisplayshortskip{4pt plus 0 pt minus 0pt}
}

\makeatletter
\renewcommand\@makefntext[1]{%
  \noindent\makebox[2pt][r]{\@makefnmark}#1}
\makeatother

\usepackage{cite}

\let\oldref=\ref                                                              
\renewcommand{\ref}[1]{\textcolor{red}{\oldref{#1}}}

\usepackage{xcolor}

\makeatother

\usepackage{babel}
\begin{document}

\title{GASP : Geometric Association with Surface Patches}

\author{Rahul Sawhney, Fuxin Li and Henrik I. Christensen\\
{\footnotesize{}\emph{\{rahul.sawhney, fli, hic\}@cc.gatech.edu}}\\
Georgia Institute of Technology
\vspace*{-8.5mm}}
\maketitle
\begin{abstract}
\,A fundamental challenge to sensory processing tasks in perception
and robotics is the problem of obtaining data associations across
views. We present a robust solution for ascertaining potentially dense
surface patch (superpixel) associations, requiring just range information.
Our approach involves decomposition of a view into regularized surface
patches. We represent them as sequences expressing geometry invariantly
over their superpixel neighborhoods, as uniquely consistent partial
orderings. We match these representations through an optimal sequence
comparison metric based on the Damerau-Levenshtein distance - enabling
robust association with quadratic complexity (in contrast to hitherto
employed joint matching formulations which are NP-complete). The approach
is able to perform under wide baselines, heavy rotations, partial
overlaps, significant occlusions and sensor noise.

The technique does not require any priors -- motion or otherwise,
and does not make restrictive assumptions on scene structure and sensor
movement. It does not require appearance -- is hence more widely applicable
than appearance reliant methods, and invulnerable to related ambiguities
such as textureless or aliased content. We present promising qualitative
and quantitative results under diverse settings, along with comparatives
with popular approaches based on range as well as RGB-D data.
\end{abstract}

\section{Introduction}

The popularity of lasers and RGB-D cameras has led to widespread use
of range and range-color images in several robotics and perception
tasks. Intrinsic to several applications, is the problem of establishing
data associations (as in \cite{neira2001JCBB,dellaert2001mcmcEM})
-- for example, in motion estimation, SLAM, SfM, loop closure / scene
detection, multi-view primitive detection and segmentation (as in
\cite{segal2009GICP,neira2001JCBB,henryRGBDMapping2010,xiao2013sun3d,whelan13deformationbasedLC,tombari2011combined,kowdle2012multiviewCoseg,lin2013holisticRgbd}).
Most 3D association solutions today, either generate sparse correspondences
on feature points, assuming a locally discriminative environment;
or use complete point clouds to indirectly associate densely based
on some form of nearest neighbors, under restricted sensor motion.

As an alternative, we present a surface patch (depth superpixel) level
association scheme. Motivated by recent trends in scene understanding
literature of utilizing superpixels due to representational compactness
and robustness to noise, we propose an analogous model for depth superpixels
- applicable over depth images acquired from generic range sensors,
where appearance / color information is not necessarily available.

Ascertaining superpixel associations without making assumptions on
sensor motion, scene appearance and/or structure, is indeed difficult
(and to our knowledge, unsolved). Superpixel decompositions vary in
each view, rendering the correspondence inexact. Besides, superpixels
are defined through (and for) homogeneity \textendash{} they are not
uniquely discriminable by design. The problem gets complicated further,
when appearance is unavailable altogether.

\begin{figure*}[t!]
\vspace*{-1.5mm}
\centering\includegraphics[max width=\linewidth, width=\linewidth,keepaspectratio]{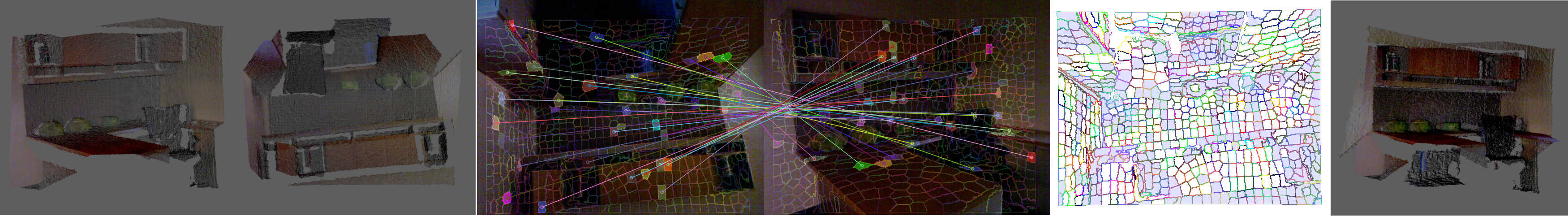}
\\[2pt]
\protect\caption{\label{fig:FirstFig} {(Best viewed in color) Point clouds from two views of a workspace scene are shown on left. The second view was captured with the sensor completely inverted ($180^{\circ}$ roll), and from a wide baseline. The two views also have significant changes in surface resolution scales, self-occlusions, and changes in yaw \& pitch. The image in centre shows a few random samples of surface patch (depth superpixel) associations between the two views, computed using our algorithm. Associated patches are connected by a line and have the same color overlay. The associations were not filtered or post-processed. The centre-right image shows the superpixel decomposition of the second view. The grey overlay over some superpixels indicates the superpixels that are not associated - these include regions which were occluded or absent in the first view. The right-most image shows the unrefined reconstruction obtained directly from the dense superpixel associations. The relative motion/transform was computed simply through corresponding 3D means of the associated superpixels.\\[0pt] }}
\end{figure*}

Nevertheless such associations are very useful - because correspondent
superpixels roughly represent the same physical 3D surface patch.
As we will show, relative scene geometry over sufficiently large neighborhoods
contains adequate discriminative information to achieve potentially
dense associations. By regularizing superpixel traits such as surface
area and smoothness \textendash{} the associations can be made to
have near co-incident 3D (centroid) localizations as well -- affording
nice sensor motion estimates even under significant change in perspectives
and scant data acquisition (\emph{Fig.} {\color{red}\emph{1}}, \emph{Table}
{\color{red}\emph{11}}). Importantly, such an approach performs equally
well in locally ambiguous (such as isomorphic or textureless) or featureless
environments. Furthermore, it can preserve localized semantics (encoded
by the superpixel labels) across views. Although not within this article's
scope, this enables more straightforward primitive level associations
and semantic transfer.

Our methodology is based on invariantly representing a (depth) superpixel
through a set of relative geometrical properties/features extracted
with respect to superpixels in its neighborhood. A uniquely consistent
ordering is utilized to sequence this set. Such a representation is
invariant to sensor's motion, and is made robust to its noise. Our
matching scheme leverages a sequence comparison metric, \emph{Restricted
Damerau Levenshtein,} \cite{damerau1964technique} -- this pertains
to family of sequence alignment algorithms, \cite{navarro2001approxStringMatching},
that have polynomial complexity, are provably optimal, and have been
in popular use for large scale sequence matching, especially in bio-informatics
community. 

As we will show, such a scheme is physically intuitive and is naturally
applicable to a geometry matching context. It is inherently robust
to heavy sensor motion, significant viewpoint differences, occlusions
and partial overlaps, and tolerant of match errors between superpixels,
due to varying decomposition across views and sensor noise.

We evaluate our approach on ground truth datasets from \cite{sturm2012RGBDbenchmark},
datasets from \cite{xiao2013sun3d}, and others collected in challenging,
yet everyday settings. The experiments are indicative of the efficacy
of the proposed approach in computing localized, dense associations.
They also indicate more robust performance than popularly used association
approaches, based on geometrical and appearance features.

\section{Related Work}

As we are unaware of literature directly addressing our problem setting,
we survey some of the relevant work in the broader scope of feature
representation and data association / matching, operating over range
and RGB-D data. We also refer to some pertinent literature in appearance
only settings. 

\emph{Point cloud association approaches} based on local 3D descriptors,
like ones used in \cite{tombari2010SHOT,rusu2009FPFH}, although useful,
can be potentially non-robust. They are hindered in settings which
are either locally homogenous or isomorphic, which is not uncommon
in everyday scenes and structures. More holistic representations have
been used for association - for example, \cite{folkesson2007mspace}
accounts for partial observability of landmarks. Plane representations
have been used in dominantly polygonal environments. \cite{taguchi2013point,dou2013planematchrgbd}
tentatively associate planes between consecutive frames based on nearest
neighbor descriptor matching and relative plane angles respectively,
before pruning them through specialized RANSAC schemes. \cite{trevor2012planarSlam,pathak2010fastPlaneregistration}
associate by assuming a physical frame to frame overlap between corresponding
planes. \emph{Registration approaches} such as \cite{segal2009GICP,stoyanov20123D-NDT}
require good initialization / restricted motion. \cite{bustos14fastrotationsearch,li20073d,olsson2006registration}
present branch and bound schemes for registration, either assuming
pure rotations or known correspondences. \cite{gelfand2005robustglobalRegist,yang2013go}
present globally optimal schemes for aligning object models, utilizing
local descriptors and interleaved ICP respectively. 

\emph{RGB-D based dense approaches} like \cite{herbst2013rgbdFlow,Kerl13VO,henry2013patch}
(and \cite{Newcombe11KFusion} which only uses depth) associate based
on flow, image warping utilizing photometric errors or ICP alignment,
to estimate motion. They afford sensor rotations, but operate under
short baseline and under the hypothesis that associations always lie
within a neighborhood epsilon. Typically, occlusions are not handled
and temporal consistency is leveraged. Such methods are suitable for
settings with constrained motion. \cite{cagniart2010probabilisticSurfaceTracking}
utilizes a patch based scheme to track deformable meshes. 

\emph{RGB-D feature based approaches }for more generic SLAM, SfM and
motion estimation applications (as in \cite{henryRGBDMapping2010,xiao2013sun3d,huang2011visual,pandey2011visbootstrapGICP})
employ sparse image features (generally SIFT, \cite{lowe2004sift})
back-projected in 3D, to ascertain frame-wise 3D correspondences.
\cite{tombari2011combined} augments the geometrical descriptor SHOT,
\cite{tombari2010SHOT}, with texture, for improved localization and
object detection.

There are \emph{higher level approaches}, which associate by leveraging
application specific constraints. \cite{ren2012rgbdLabelling} utilizes
aggregation of densely sampled point features at superpixel levels
for RGB-D object detection and recognition. \cite{mivcuvsik2010superpixelStereo}
utilizes (color only) superpixel associations to reconstruct piecewise
planar scenes under known extrinsics. It assumes similar sensor orientations
and imposes restrictions on possible plane orientations. Stereo literature
like \cite{bleyer2010surfaceStereoSoft,sinha14stereoLocalPlaneSweeps}
ascertain disparity maps by associating surfaces / planes relying
on short baselines, similar sensor orientations, and discriminative
appearance or local features. \cite{kowdle2012multiviewCoseg} utilizes
planar stereo reconstructs to segment fully observable foreground
from multiple views. \cite{bodis14piecewiseplanarmodelling,Gallup10peicwiseRecon}
operate upon SfM point clouds (reconstructed a priori) to reason about
visibility and association of planar primitives from multiple views.
\cite{eshet2013DCSH,barnes2010genPatchMatch,hacohen2011nonrigidDenseCorrespond}
utilize appearance similarity at patch levels to build a dictionary
and associate in the nearest neighbor sense. They find use in image
enhancement, and matching scenes with similar appearance.

\emph{Matching methodologies}, apart from typically assuming availability
of discriminative local features, quite often also rely on motion
and/or visibility priors - \cite{neira2001JCBB} for example, performs
exhaustive search over all possible permutations of joint associations
(exponential complexity), and attains tractability through priors.
Similarly, intractable joint probabilistic formulations used in \cite{thrun1998probabilistic,dellaert2001mcmcEM},
attain feasibility through priors. Graph techniques have been used
to ascertain jointly consistent feature matches. \cite{conte2004thirtyyrsGraph}
presents a good overview -- The approach is to represent features
as nodes, with the relative constraints between them as graph edges.
An edge preserving mapping between nodes of such graphs is then computed,
as either a subgraph isomorphism, or relaxed to inexact graph homomorphism,
or as bipartite graph matching problem with non-linear constraints
(say, when edges are distances). All of the above formulations are
NP-complete and become quickly intractable, especially in absence
of priors. Exact matching formulations have been mostly limited to
sparse 2D scenarios, as in \cite{torresani2013dualdecomp,bailey2000data,caelli2005graphicalmodelsGraphmatch},
involving a rather limited number of nodes. \cite{bailey2000data}
utilizes maximum common subgraph formulation to associate sparse 2D
laser scans. \cite{olson2005singleclustSpectralGraph} approximates
a dominant solution through eigenanalysis of the graph adjacency matrix.
\cite{cho2012progressiveGraphmatch,Wang2013densitymaximGraphMatch}
present recent approximate graph based solutions for ascertaining
image feature matches -- \cite{cho2012progressiveGraphmatch} progressively
improves skeletal graphs, while \cite{Wang2013densitymaximGraphMatch}
employs a density maximization scheme.

\section{Methodology}

We first segment/decompose a view into regularized depth superpixels/surface
patches (\emph{Sec.} \emph{\ref{sub:Depth-Superpixel-Decomposition}}).
Each superpixel (or the ones of interest) is then expressed through
a set of geometric features/relationships arising from all the superpixels
in its neighborhood. Each feature in the set corresponds to a superpixel
in the considered neighborhood, and is defined through patch level
relative geometrical properties expressed invariantly (\emph{\ensuremath{S}ec}.
\emph{\ref{sub:Representing-Rigid-Geometry}}). The ascertained feature
set, thus, jointly represents all geometrical features of interest
in the neighborhood. Finer or coarser geometrical detail can be captured
by adjusting the granularity of the superpixel decomposition. Similarly,
more global (or local) geometry can be represented by considering
larger (or smaller) neighborhoods. 

Such a representation captures invariant 3D geometry effectively.
It does not require assumptions of the scene structure, such as piecewise
or dominant planarity. It is also discriminative enough to disambiguate
in difficult settings such as ones with duplicate or locally isomorphic
content (\emph{Fig.} \emph{\ref{fig:qual}}).

The feature set of a superpixel is then arranged as a sequence by
enforcing an ordering over them (\emph{\ensuremath{S}ec}. \emph{\ref{sub:Ordering}}).
The motivation for sequencing is to induce a partial order which remains
invariant across views. This is required so that feature sets from
different views can be correctly matched\emph{. }

Our matching scheme (\emph{Sec. \ref{sub:Matching}}) utilizes edit
distance based sequence comparisons (\cite{navarro2001approxStringMatching}),
specifically the Restricted Damerau Levenshtein distance metric (\cite{damerau1964technique}).
The edit distance between two sequences of arbitrary length can be
optimally evaluated in quadratic time, and is directly indicative
of their dissimilarity. In our context, a feature sequence expresses
neighborhood geometry about a given superpixel - with each of its
features exclusively capturing geometric information corresponding
to a neighboring surface patch. Comparisons between two such sequences,
thus, gives us powerful means to ascertain the amount of geometrical
mismatch between the two considered neighborhoods%
\footnote{By tractable comparisons, we can now essentially match geometry between
3D neighborhoods in as globalized (or localized) manner, as desired.
This is especially useful for cases where geometry about localized/small
neighborhoods is not discriminative enough for making associations;
large/global neighborhoods need to be considered to disambiguate then.%
}. The approach is also inherently robust - as scenarios with partial
view overlaps, occlusions and self-occlusions are naturally afforded
through the edit operations, and sensor noises can be intuitively
accounted for while matching individual features in the sequences
(\emph{Table }$\ref{tab:quant}$ quantifies GASP's performance with
increasing baselines, perspective changes and non-overlapping content).

We specify the superpixel decomposition of given a given view of a
scene as $S=\{s_{i}\}_{i=1}^{N}$. We denote $\mu\,$ ($\in S$) as
the superpixel currently under consideration. Similarly, another view
of the scene will have a decomposition $S'=\{s'_{j}\}_{j=1}^{N'}$.
$\mu'$ would denote a superpixel from $S'$ currently being considered
for possible correspondence with $\mu$. $\aleph_{\mu}$ indicates
the set of nearest superpixels in $\mu$'s 3D neighborhood, with $|\aleph_{\mu}|$
indicating the cardinality of the set. $\alpha$ refers to an arbitrary
superpixel in $\mu$'s neighborhood ($\alpha\in\aleph_{\mu}$).

\subsection{\label{sub:Depth-Superpixel-Decomposition}Depth Superpixel Decomposition}

Our segmentation approach essentially involves decomposing each contiguous,
3D surface (not necessarily planar), into compact (not necessarily
small) smooth patches of similar surface area. The extracted superpixels
thus maintain consistency across views and are uniform in 3D. We will
only present a procedural overview here. A range image is first segmented
into a set of contiguous components, where each component respects
3D surface edges and lies entirely on a smooth surface. This is done
by comparing adjacent points for 3D connectivity, normal angles, depth
disparity and utilizing curvature. Small components on/near surface
edges (due to unavoidable, irregular smoothing of normals) are initially
ignored and their constituent points are added through post processing.
Each remaining component/surface is then decomposed into patches of
similar area. This is done using K-Means with initialization seeds
spread uniformly across the 3D surface (similar to \cite{arthur2007kmeanspp}).
Post segmentation, the remaining depth image points, and superpixels
below a minimum point size are agglomerated back into the regular
superpixels based on proximity and matching normals. Graph representation
and operations (edge contraction, vertex splitting) are utilized for
ease and ensuring superpixel contiguity. The 3D mean of a superpixel\textquoteright s
constituent points served as a sufficiently good location estimate.
Similarly, the superpixel normal is taken as the point normals' mean.
Due to surface area regularization, parts of a scene closer to the
sensor would have larger superpixels (more pixels) than the ones farther
away (\emph{Fig.} \emph{\ref{fig:Ordering}}). Similarly, a view from
farther away would have more superpixels, as it covers more of the
scene area (\emph{Fig.} \emph{\ref{fig:qual:umdbed}}).

\subsection{\label{sub:Representing-Rigid-Geometry}Representing Geometry}

We utilize $\aleph_{\mu}$, to express $\mu$ through a set of transformation
invariant geometrical features, $Q_{\mu}=\{q_{\mu}^{\alpha}\}_{\forall\alpha\in\aleph_{\mu}}$.
Each superpixel, $\alpha$, in the neighborhood, $\aleph_{\mu}$,
contributes geometric information, $q_{\mu}^{\alpha}$ , and helps
capture the geometry in $\mu$\textquoteright s 3D neighborhood. Let
$\hat{n}_{\mu}$ indicate $\mu$\textquoteright s surface normal,
and $l_{\mu}$ indicate its 3D location. $\overrightarrow{r}{}_{\mu}^{\alpha}=l_{\alpha}-l_{\mu}$
, would indicate the relative displacement of the superpixel; $\hat{r}{}_{\mu}^{\alpha}$
and $||\overrightarrow{r}{}_{\mu}^{\alpha}||$ would indicate its
direction and magnitude respectively. Evidently, the quantities $\hat{n}_{\mu},\,\hat{n}_{\alpha},\,\overrightarrow{r}{}_{\mu}^{\alpha}$
depend on the reference frame. In order to make them invariant to
the sensor pose, we express them in a coordinate frame derived from
superpixels $\mu$ and $\alpha$ themselves. An orthonormal co-ordinate
frame can be derived from $\hat{n}_{\mu}$ and $\overrightarrow{r}{}_{\mu}^{\alpha}$
as follows :

\begin{subequations}
\begin{gather}
\hat{u}_{\mu}^{\alpha}=\hat{r}{}_{\mu}^{\alpha}\\
\hat{v}_{\mu}^{\alpha}=\frac{\hat{n}_{\mu}-\left(\hat{n}_{\mu}\cdot\hat{r}{}_{\mu}^{\alpha}\right)\hat{r}{}_{\mu}^{\alpha}}{\left\Vert \hat{n}_{\mu}-\left(\hat{n}_{\mu}\cdot\hat{r}{}_{\mu}^{\alpha}\right)\hat{r}{}_{\mu}^{\alpha}\right\Vert }\\
\hat{w}_{\mu}^{\alpha}=\frac{\hat{u}_{\mu}^{\alpha}\times\hat{v}_{\mu}^{\alpha}}{\left\Vert \hat{u}_{\mu}^{\alpha}\times\hat{v}_{\mu}^{\alpha}\right\Vert }
\end{gather}
\label{eq:Coord}
\end{subequations}

\vspace*{-12pt}

where $\hat{u}_{\mu}^{\alpha},\,\hat{v}_{\mu}^{\alpha},\, w_{\mu}^{\alpha}$
form an orthonormal basis. This basis is almost never degenerate,
as $\hat{n}_{\mu}$~and~$\overrightarrow{r}{}_{\mu}^{\alpha}$ are
rarely co-linear. $\hat{n}_{\alpha}$ can now be expressed in this
local frame through the projection components $[\hat{n}_{\alpha}\cdot\hat{u}_{\mu}^{\alpha},\,\hat{n}_{\alpha}\cdot\hat{v}_{\mu}^{\alpha},\,\hat{n}_{\alpha}\cdot\hat{w}_{\mu}^{\alpha}]^{T}$.
For two given superpixels, $\mu$ and $\alpha$, these components
would remain independent of the sensor viewing frame. Additional pieces
of relative, invariant information can be extracted through $\hat{n}_{\mu},\hat{\, n}_{\alpha}$
and $\overrightarrow{r}{}_{\mu}^{\alpha}$ as follows:

\begin{subequations}
\begin{gather}
\theta_{\alpha,\mu}=cos^{-1}\left(\hat{n}_{\mu}\cdot\hat{n}_{\alpha}\right)\\
\theta_{r,\mu}=cos^{-1}\left(\hat{r}{}_{\mu}^{\alpha}\cdot\hat{n}_{\mu}\right)\\
\theta_{r,\alpha}=cos^{-1}\left(\hat{r}{}_{\mu}^{\alpha}\cdot\hat{n}_{\alpha}\right)
\end{gather}
\label{eq:RelThetas}
\end{subequations}

\vspace*{-12pt}

$q_{\mu}^{\alpha}$ can now be expressed as a feature vector constituting
of seven relative, invariant elements. We have then :

\begin{multline} 
q_{\mu}^{\alpha}=\left[\right.||\overrightarrow{r}{}_{\mu}^{\alpha}||\cdot sgn(\hat{n}_{\alpha}\cdot\hat{u}_{\mu}^{\alpha}),~||\overrightarrow{r}{}_{\mu}^{\alpha}||\cdot sgn(\hat{n}_{\alpha}\cdot\hat{v}_{\mu}^{\alpha}),\dots \\
\dots||\overrightarrow{r}{}_{\mu}^{\alpha}||\cdot sgn(\hat{n}_{\alpha}\cdot\hat{w}_{\mu}^{\alpha}),~||\overrightarrow{r}{}_{\mu}^{\alpha}||,~\theta_{\alpha,\mu},~\theta_{r,\mu},~\theta_{r,\alpha}\left.\right]^{T}  
\end{multline}

where $||\overrightarrow{r}{}_{\mu}^{\alpha}||$ is included for additional
redundancy. We utilize the signs of $\hat{n}_{\alpha}$'s projection
components (with $sgn(.)\in\{-1,0,1\}$), as their actual values tend
to be noisy. A proper approach is to utilize an epsilon-insensitive
signum function, for example $sgn_{e}(.)$ rather than $sgn(.)$ -
it is defined as zero either when $1)$ the angle between $\hat{n}_{\alpha}$
and the respective basis vector $\not\in[e_{\theta},\, PI-e_{\theta}]$;
or in the uncommon case of degenerate basis when $2)$ $\hat{n}_{\mu}$
is co-linear with $\overrightarrow{r}{}_{\mu}^{\alpha}$ (that is,
when $\theta_{r,\mu}\not\in[e_{\theta},\, PI-e_{\theta}]$). $e_{\theta}$
is the allowable angular noise tolerance. The component signs are
then scaled by $||\overrightarrow{r}{}_{\mu}^{\alpha}||$ in order
to incorporate signed distance information between $\mu$ and $\alpha$.
The feature, $q_{\mu}^{\alpha}$, is thus expressed stably in presence
of noises. It captures relative information between $\mu$ and the
neighborhood superpixel $\alpha$ \textendash{} the relative pose,
distance, orientation and bearings. Understandably, the elements of
$q_{\mu}^{\alpha}$ would be affected by noise. However, our matching
methodology is robust to it, explicitly accounts for it ($Sec.$ \emph{\ref{sub:Matching}},
deviation thresholds). The joint feature set $Q_{\mu}=\{q_{\mu}^{\alpha}\}_{\forall\alpha\in\aleph_{\mu}}$,
constituting of relative feature vectors from all superpixels in $\aleph_{\mu}$,
thus, essentially expresses the geometry in $\mu$\textquoteright s
neighborhood invariantly.

\subsection{\label{sub:Ordering}Ordering}

\begin{figure}[t!]
\vspace*{-3mm}
\includegraphics[width=1\columnwidth]{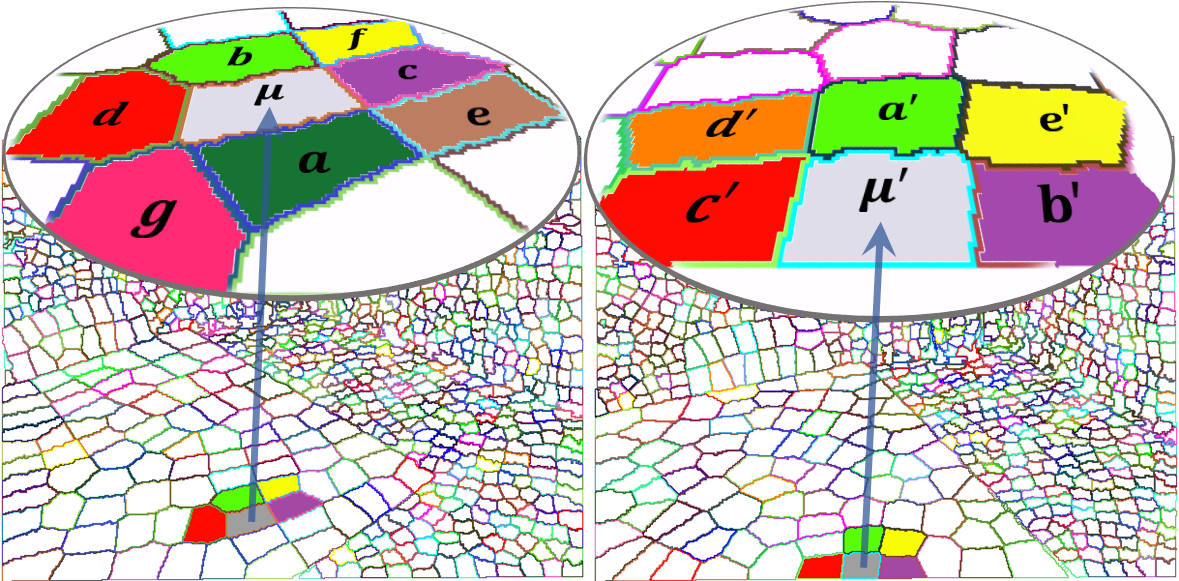}\\[-8pt]
\protect\caption{\label{fig:Ordering}Illustrative consistent orderings of immediate neighborhoods of associating superpixels, $\mu$ and $\mu'$ are shown. The orderings are indicated by alphabetical progression of the marked neighboring superpixels. The matching pairs of superpixels are shown on the table, and share a common color. The orderings are consistent as the sets of corresponding neighborhood superpixels \{\textbf{\textcolor{black}{\emph{b}}}, \textbf{\textcolor{amethyst}{\emph{c}}}, \textbf{\textcolor{candyapplered}{\emph{d}}}, \textbf{\textcolor{aureolin}{\emph{f}}}\} and  \{\textbf{\textcolor{black}{\emph{a'}}}, \textbf{\textcolor{amethyst}{\emph{b'}}}, \textbf{\textcolor{candyapplered}{\emph{c'}}}, \textbf{\textcolor{aureolin}{\emph{e'}}}\}, as indicated by their increasing  alphabetic order, arise identically in the orderings.}
\end{figure}Once the geometric feature set has been determined, a partial ordering
needs to be imposed on its elements to obtain a feature sequence -
as our matching scheme leverages a distance metric based on sequence
aligning comparisons. An ordering over $\mu$\textquoteright s neighborhood
$\aleph_{\mu}$, can be denoted as $\overset{\leftharpoonup}{O}_{\mu}=\left\langle a,b,c\dots\alpha\dots\right\rangle ,\,\forall\alpha\in\aleph_{\mu}$
\textendash{} where $\left\langle a,b,c\dots\right\rangle $ indicates
the ordered sequence of superpixels. This is used to order the joint
feature set $Q_{\mu}$ as $\overset{\leftharpoonup}{Q_{\mu}}=\left\langle q_{\mu}^{a},q_{\mu}^{b},q_{\mu}^{c}\dots q_{\mu}^{\alpha}\dots\right\rangle $.
These ordered sequences are subsequently utilized to ascertain a potential
association between two given superpixels, $\mu\,\&\,\mu'$ from different
views. The approach is to devise an ordering scheme such that two
superpixel orderings, $\overset{\leftharpoonup}{O_{\mu}\,}\&\,\overset{\leftharpoonup}{O_{\mu'}}$
(over $\aleph_{\mu}\,\&\,\aleph_{\mu'}$) are both partially ordered
by (with respect to) their matching subsequences - these subsequences
would be the identically ordered (sub-)sets of corresponding superpixels
in neighborhoods of $\mu$ and $\mu^{'}$ respectively. To put it
simply, \emph{the order of the correctly corresponding superpixels
in the neighborhoods} $\aleph_{\mu}$ \emph{and} $\aleph_{\mu'}$
\emph{respectively}, \emph{should remain invariant in the respective
orderings, }$\overset{\leftharpoonup}{O_{\mu}}$ and $\overset{\leftharpoonup}{O_{\mu'}}$\emph{
}($\overset{\leftharpoonup}{O_{\mu}}$ and $\overset{\leftharpoonup}{O_{\mu'}}$
should be consistent). \emph{Fig.} \ref{fig:Ordering} illustrates
consistent orderings, $\overset{\leftharpoonup}{O}_{\mu}=<a,b,c,d,e,f,g>$
and $\overset{\leftharpoonup}{O}_{\mu'}=<a',b',c',d',e'>$, over immediate
neighborhoods of two correctly associating superpixels $\mu$ and
$\mu'$. These orderings are \emph{mutually} consistent as the correct
correspondences in the neighborhoods form subsequences -- $<b,c,d,f>$
and $<a',b',c',e'>$ arise in identical order in $\overset{\leftharpoonup}{O}_{\mu}$
and $\overset{\leftharpoonup}{O}_{\mu'}$ respectively.

To achieve ordering consistency, we utilize $Q_{\mu}$ itself which
already constitutes of a superpixel-wise set of invariant geometric
features relative to $\mu$. In effect, $\overset{\leftharpoonup}{Q_{\mu}}$
is simply ascertained through a robust sorting operation over $Q_{\mu}$'s
elements ($\{q_{\mu}^{\alpha}\}$, which are invariant). Note that
since this ordering is derived from geometry with respect to $\mu$
(\emph{Eqs. \ref{eq:Coord}, \ref{eq:RelThetas}}), it will not, in
general, be consistent with an ordering over an arbitrary superpixel
from $S'$. It will only be consistent with an ordering, say $\overset{\leftharpoonup}{O}_{\mu'}$,
defined about some superpixel $\mu'$ in $S'$, which has similar
relative geometry in its neighborhood as $\mu$ \textendash{} which
is precisely the objective%
\footnote{$\overset{\leftharpoonup}{O}_{\mu}$ will, in fact, be quite inconsistent
with an ordering about a non-corresponding superpixel in $S'$ and
hence, as a consequence of mutually inconsistent orderings, will result
in rather poor match distance. %
}. Thus $\overset{\leftharpoonup}{Q_{\mu}}=Sort(Q_{\mu},\, e_{r},\, e_{\theta})$;
with the sorter conducting pairwise comparisons between $Q_{\mu}$'s
constitutent features. The second dimension (of the two features being
compared) is only used if the first dimension is equivalent, the third
dimension is only used if the first two are equivalent, and so forth.
Equivalence is defined as the values being within epsilon tolerances
of each other, to affect resolution, and account for noise and finite
precision numerical errors. A distance tolerance, $e_{r}$ is used,
along with the afore-utlilized angular tolerance, $e_{\theta}.$ In
our experiments over noisy Kinect data, $e_{r}=.02$ metres and $e_{\theta}=\nicefrac{5\pi}{180}$
radians, worked well.

\begin{algorithm}[t!]
\vspace*{-3mm}
\caption{ Compare Geometric Sequences}\label{Algomatch} 
\fontsize{8pt}{10pt}\selectfont
\begin{algorithmic}[1] 
\NoThen
\NoDo
\vspace{1mm}
\Function{\emph{CompareRDL}}{ $\overset{\leftharpoonup}{Q_{\mu}}, \overset{\leftharpoonup}{Q_{{\mu}'}}$ } 
\hspace{3mm}{\textbf{Returns } $D_{RDL}(\mu,\mu')$}    
\State $insert \gets 1;~delete \gets 1;$
\State $replace \gets \infty;~switch \gets 0;$
\State $L_\mu \gets |\overset{\leftharpoonup}{Q_{\mu}}|;~L_{{\mu}'} \gets |\overset{\leftharpoonup}{Q_{{\mu}'}}|;$
\State $tab(1,1) \gets 0;$
\ForAll{$i \in [2,L_\mu]$}
\State $tab[i,1] \gets tab[i-1,1]+insert$
\EndFor
\ForAll{$j \in [2,L_{{\mu}'}]$}
\State $tab[1,j] \gets tab[1,j-1]+delete$
\EndFor
\ForAll{$j \in [2,L_{{\mu}'}]$}
\ForAll{$i \in [2,L_\mu]$}
\If{$Match (\overset{\leftharpoonup}{Q_{\mu}}(i),\overset{\leftharpoonup}{Q_{{\mu}'}}(j))$}
\State $~~substitute \gets 0$
\Else $~~substitute \gets replace$
\EndIf
\State $tab[i,j] \gets min\Biggl\{\begin{array}{c} tab[i-1,j]+insert,\\ tab[i,j-1]+delete,\\ tab[i,j-1]+substitute\end{array}\Biggr\}$
\If{$i>2~\&~j>2~\&~Match (\overset{\leftharpoonup}{Q_{\mu}}(i-1),\overset{\leftharpoonup}{Q_{{\mu}'}}(j-1))$}
\State $~~tab[i,j] \gets min(tab[i,j],tab[i-2,j-2]+switch)$
\EndIf
\EndFor
\EndFor \\~~
\Return{$tab[L_{\mu},L_{\mu'}]$} 
\EndFunction 
\end{algorithmic}
\hfill
\begin{algorithmic}[1] 
\NoThen
\NoDo
\Function{\emph{Match}}{$~q^{\beta},~q^{\gamma}~$} 
\hspace{3mm}{\textbf{Returns } $True~or~False$}    
\State $r_{dev} \gets UserDefined;~\theta_{dev} \gets UserDefined;$
\State $q_{noise} \gets [\theta_{dev},\theta_{dev},\theta_{dev},r_{dev},r_{dev},r_{dev},r_{dev}]$ 
\State $\Delta q \gets abs(q^{\beta} - q^{\gamma})$
\ForAll{$t \in [1,7]$}
\If{$ \Delta q[t] > q_{noise}[t]$}
\State \textbf{Return} $False$
\EndIf
\EndFor\\~~
\Return{$True$}
\EndFunction
\end{algorithmic}
\end{algorithm}

\subsection{\label{sub:Matching}Matching}

A pair of superpixels, $\mu$ and $\mu'$, can now be compared for
geometric correspondence using their respective ordered feature sequences
$\overset{\leftharpoonup}{Q_{\mu}}$ and $\overset{\leftharpoonup}{Q_{\mu'}}$.
We utilize a sequence matching scheme based on \cite{damerau1964technique},
to ascertain/quantify the dissimilarity between the sequences in the
form of edit distances. Edit distance based algorithms, \cite{navarro2001approxStringMatching},
operate by editing one sequence into another. By utilizing efficient
dynamic programming, they progressively compare two elements at a
time, one from each sequence. If the elements match up, the next pair
of elements is considered \textendash{} else element in one of the
sequences is edited first at a cost (typically by inserting, deleting
or replacing it), before resuming the comparisons. Sequences which
have matching elements will result in lower edit distances than ones
which do not. Additionally, sequences which have matching elements
in the same order will result in lower edit distances than ones which
do not. The computed distances for algorithms such as \cite{damerau1964technique}
are optimal with respect to the specified editing costs.

Two features, $q_{\mu}^{\alpha}$ and $q_{\mu'}^{\alpha'}$, from
the respective sequences $\overset{\leftharpoonup}{Q_{\mu}}$ and
$\overset{\leftharpoonup}{Q_{\mu'}}$, will match up when relative
geometry of patch $\alpha$ with respect to $\mu$ , is the same as
the relative geometry of $\alpha'$ with respect to $\mu'$ ($q_{\mu}^{\alpha}$
and $q_{\mu'}^{\alpha'}$ would then hold approximately same values).
If $\mu$ and $\mu'$ have the same relative geometry in their neighborhoods,
the sequences on the whole will naturally match, and will not require
many edit operations \textendash{} resulting in low edit distances;
else the distances will be high. Additionally, by devising an ordering
utilizing the unique relative features themselves, two sequences which
do not capture similar geometry will match up badly, because their
ordering will differ significantly. We utilize the \emph{Restricted
Damerau\textendash Levenshtein} (RDL, \cite{damerau1964technique})
algorithm for ascertaining sequence disparity. In contrast to the
popular Levenshtein algorithm (\cite{navarro2001approxStringMatching}),
which allows insert, delete and replace operations over sequence elements,
RDL allows the additional operation of transposition of adjacent elements%
\footnote{Assuming all edit weights to be unity, the Levenshtein distance between
string sequences, $ABCD~\&~BAC$, is $3$, while the Restricted Damerau-Levenshtein
distance is $2$ \textendash{} due to an aligning transposition.%
} , and has the additional constraint that each subsequence can be
edited only once. The edit operations' costs can be set arbitrarily
to suit a use case, and to achieve a desired resolution. Comparing
two sequences through RDL is a symmetric operation, and sequences
of different sizes can be compared. 

These properties suit our needs nicely. Superpixels in the neighborhood
$\aleph_{\mu}$ may not be present in $\aleph_{\mu^{'}}$ and vice
versa. This would be because of partial overlap of content between
views, and because of occlusions. The operations of insert and delete
will basically edit such non-matching features from the sequences.
The ability to transpose adjacent features accounts for slight errors
in the sequence orderings%
\footnote{Our experiments indicated that, in less noisy settings/good datasets,
the transposition operation could be optionally disabled without significant
impact on association accuracies, due to the use of epsilon tolerances.%
}. Replacement in this context, being physically meaningless, is disabled.
Also, the restriction that each subsequence can be altered only once,
prevents any re-edits over incumbent feature alignments. Insertion,
deletion are symmetric operations and we nominally set their cost
to unity. Transposition cost is set to zero, as it only occurs only
due to slight ordering inconsistencies. 

\begin{figure}[t!]
\vspace*{-2mm}
\includegraphics[width=1\columnwidth]{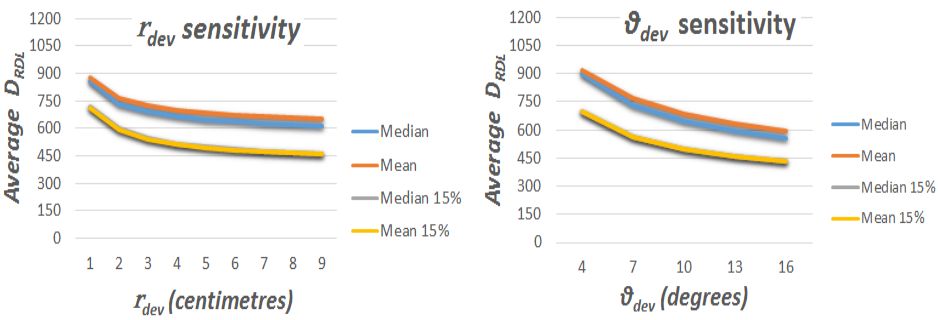}\\[-9pt]
\protect\caption{\label{fig:SensPlot}Impact of varying match thresholds $r_{dev}~\&~\theta_{dev}$ on averaged $D_{\mu_{RDL}}$ is shown. Default $r_{dev}$, $\theta_{dev}$ were  $5$ cm and $10^{\circ}$. Mean and Median edit distances, over  all associations, and over top $15\%$ are shown.}
\end{figure}Obtaining a match between two given features, $q_{\mu}^{\alpha}$
and $q_{\mu'}^{\alpha'}$, while comparing $\overset{\leftharpoonup}{Q_{\mu}}$
and $\overset{\leftharpoonup}{Q_{\mu'}}$ is easy. Basically, a match
is established when the respective components of the two features
lie within some acceptable range of each other. Two simple, intuitive
thresholds \textendash{} one for allowable angular deviation, $\theta_{dev}$
and the other for allowable distance deviation, $r_{dev}$ \textendash{}
are utilized. These thresholds account for noise and allowable slack
in elements of $q_{\mu}^{\alpha}$ and $q_{\mu'}^{\alpha'}$. \emph{Fig.
\ref{fig:SensPlot}} plots the effect of varying them on (average)
edit distances. Also, more precise, co-incident localizations can
be achieved by considering smaller values, and a more granular superpixel
discretization (vice versa is applicable too). 

Two superpixel sequences from different views corresponding to the
same 3D location, should have zero edit costs \textendash{} assuming
complete overlap of neighborhoods, no occlusions and consistent segmentations.
In practice, correctly associated superpixels (their sequences) would
still have some edit costs due to partially overlapping neighborhood
geometry, occluded regions, and inexact superpixel decomposition across
views. Desirably, these absent, occluded, or mismatched superpixel
features would be edited out. Incorrect associations will have significantly
higher edit distances, as a consequence of dissimilar neighborhood
geometry. \emph{Alg. }\emph{\ref{Algomatch}} specifies the matching
algorithm. It returns the net edit cost, $D_{RDL}(\mu,\mu')$, between
the feature sequences of $\mu$ and $\mu'$, being considered for
association. For clarity, embellishments for memory and computational
efficiency have been left out. Note that $D_{RDL}(\mu,\mu')\equiv D_{RDL}(\mu',\mu)$.

\subsection{\label{sub:Ascertaining-Associations}Ascertaining Associations}

The best potential, putative association for a superpixel, $\mu\in S$
, would be the superpixel in $S'$ whose neighborhood geometry matches
most with $\mu$'s neighborhood - one whose feature sequence gives
the lowest edit distance with $\overset{\leftharpoonup}{Q_{\mu}}$. 

\begin{equation}
D_{\mu_{RDL}}=min_{\forall\alpha'\in S'}(D_{RDL}(\mu,\alpha'))
\end{equation}

where $D_{\mu_{RDL}}$ indicates the edit distance from the best association
in $S'$, $\mu_{best}'\equiv argmin_{\forall\alpha'\in S'}(D_{RDL}(\mu,\alpha'))$.
$D_{\mu_{RDL}}$is basically indicative of the amount of rigid geometry
mismatch between the neighborhoods of $\mu$ and its best putative
association; and can hence be utilized to ascertain whether the putative
association is considered correct. Normalized edit distances are used
for this purpose%
\footnote{With pertinent application specific adjustments, normalized edit distances
are amenable to a probabilistic interpretation as well.%
}. For view to view matching, all superpixels in a view can be made
to use equal size neighborhoods (that is, the set of nearest superpixels
in 3D) ; thus $|\aleph_{\mu}|=k_{S},~\forall\mu\in S$ and $|\aleph_{\mu'}|=k_{S'},~\forall\mu'\in S'$.
$D_{RDL}(\mu,\mu')$, for any $\mu$ and $\mu'$, can thus have a
maximum value of $(k_{S}+k_{S'})$, assuming unit costs for insert/delete
operations. Normalized edit distance is then obtained as :

\begin{equation}
\hat{D}_{\mu_{RDL}}=\frac{min_{\forall\alpha'\in S'}(D_{RDL}(\mu,\alpha'))}{(k_{S}+k_{S'})}
\end{equation}

$D_{\mu_{RDL}}/\hat{D}_{\mu_{RDL}}$ are dependable measures of association
quality. A putative association for a given superpixel, $\mu$, is
considered correct in the geometric sense, if $\hat{D}_{\mu_{RDL}}$
is not more than a given normal gating value, $\lambda\in\left[0,1\right]$
($\hat{D}_{\mu_{RDL}}\leq\lambda$ for association). A lower $\lambda$
would result in more confident and localized associations, while denser
but possibly coarser associations would arise at higher $\lambda$
gatings%
\footnote{Alternatively, or when more accurate metric transforms is the end
objective, a dense set of putative associations could be first obtained
using a high gating; these could be subsequently filtered on the basis
of 3D rigid transformation consistency, using a scheme such as RANSAC
(\emph{Sec. \ref{sec:results}}). Similarly, for a semantic transfer/segmentation
task, associations obtained with a high gating could be smoothed out,
in a framework such as Conditional Random Fields.%
} (depending on a scene\textquoteright s geometrical ambiguity, considered
neighborhood sizes and match deviation thresholds).

\begin{table}[t!]
\scriptsize
\setlength{\belowrulesep}{1pt} 
\setlength{\aboverulesep}{1pt}
\begin{tabular*}{1\columnwidth}{l|p{7.5mm}|p{7.5mm}|p{7.5mm}|p{7.5mm}}
\toprule
\textbf{\emph{\# Feature means queried - C}} & $25$ & $50$ & $75$ & $100$\tabularnewline
\midrule
\textbf{\emph{$\mathbf{\mu_{best}'}$~found (\% Avg.,~$\mathbf{\lambda=.5}$)}}  & $71.5$ & $86.8$ & $94.5$ & $98.9$\tabularnewline
\bottomrule 
\end{tabular*}\\[2pt]
\protect\caption{\label{tab:kdquery}Averaged \% of best associations with increasing query sizes.}
\end{table}

Obtaining a putative association, when comparing with all superpixels
in $S'$, would have a worst case complexity of $O(||S'|k_{S}k_{S'}|)$.
Although the cubic complexity is tractable, significant further improvements
are possible. Some discriminative information can be leveraged from
the feature set, $Q_{\mu}$'s mean, $\overline{Q}_{\mu}$. If two
superpixels, $\mu$ and $\mu'$ form a correct correspondence, their
respective feature set averages, $\overline{Q}_{\mu}$ and $\overline{Q}_{\mu'}$
would be close to each other. To find the putative match for $\mu$
in $S'$, we therefore build a KD-tree over feature set means (normalized)
of all superpixels in $S'$, and search/query for $C$ of the nearest
neighboring feature set means to $\overline{Q}_{\mu}$. The putative
association, and subsequently a possible correct association for $\mu$,
is then ascertained from the superpixels corresponding to these queried
feature means rather than considering all superpixels in $S'$. This
brings down the complexity of ascertaining a putative association
to quadratic \textendash{} $O(Ck_{S}k_{S'})$, where $C$ is the constant
number of queries, with $C\ll S'$. \emph{Table} \emph{\ref{tab:kdquery}}
indicates the average percentage of best associations ($\mu_{best}'$)
found, as a function of query size, $C$ \textendash{} at least an
order of magnitude reduction in computations is achieved, without
any significant impact on association accuracies. An early termination
criteria in \emph{Alg.} \emph{\ref{Algomatch}} gives another significant
improvement. Since associations with normalized distances above $\lambda$
are anyways ignored, \emph{Alg.} \emph{\ref{Algomatch}} can be terminated
prematurely as soon as the edit costs exceed $\lambda(k_{S}+k_{S'})$.
This is generally the case for a majority of potential associations
in $S'$, and results in significant gains in practice. Further gains
are possible, like screening of possible associations before computing
$D_{RDL}$, utilizing progressive rigid transform estimates. Also
note that the associations are computable in parallel - such kinds
of efficiency gains is a subject of ensuing work.\begin{figure*}[t!]
\newcommand{\sepspacebelow}{-.3em}
\newcommand{\sepspaceabove}{0em}
\newcommand{\spacebelowcaption}{-.8em} 
\newcommand{\spaceabovecaption}{.3em} 
\centering
\begin{subfigure}[h]{1\linewidth}
{\vspace*{-4mm}
\caption{\label{fig:qual:house} {Clutter scene from \cite{sturm2012RGBDbenchmark}. It has complex geometry and self-occlusions. The views have significant change in perspective, surface resolution scales.~~~~~~~~~~~~~~~~~~~} \vspace*{\spacebelowcaption}}
\includegraphics[max width=\linewidth, width=\linewidth,keepaspectratio]{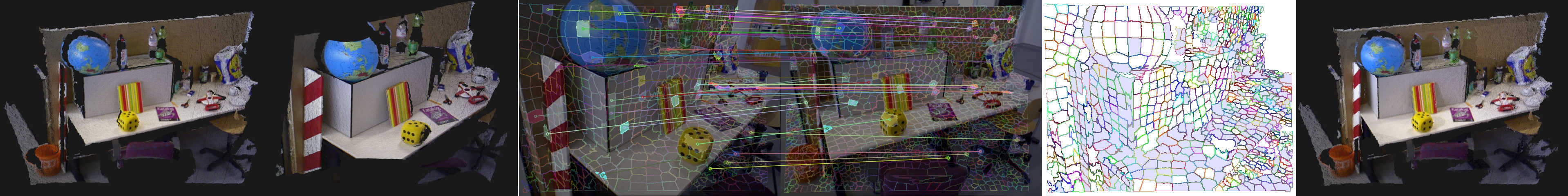}
}\end{subfigure}\hfill
\vspace*{\spaceabovecaption}
\begin{subfigure}[h]{1\linewidth}
{\caption{\label{fig:qual:symmetry} {Scene with multiple primitive instances in a near symmetrical setting (points outside the sofa setting volume were clipped off). The views have a full roll inversion, and changes in pitch and yaw as well.}\vspace*{\spacebelowcaption} }
\includegraphics[max width=\linewidth, width=\linewidth,keepaspectratio]{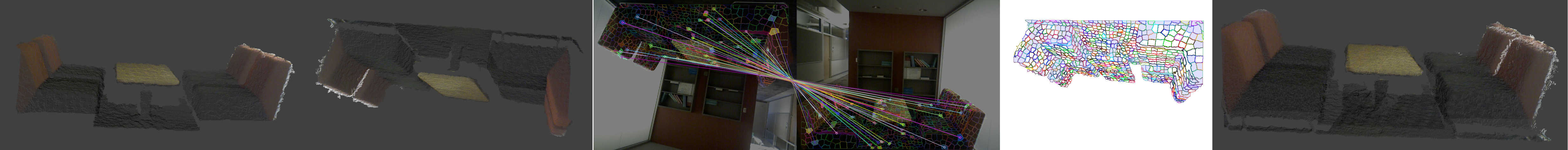}
}\end{subfigure}\hfill
\vspace*{\spaceabovecaption}
\begin{subfigure}[h]{1\linewidth}
{\caption{\label{fig:qual:sparse} {Results over a scene with sparse structure and duplicate primitives. The views have a full roll inversion. An occluding body was introduced in the second view - the superpixels pertaining to it will not get matched (grey overlay).} \vspace*{\spacebelowcaption} }
\includegraphics[max width=\linewidth, width=\linewidth,keepaspectratio]{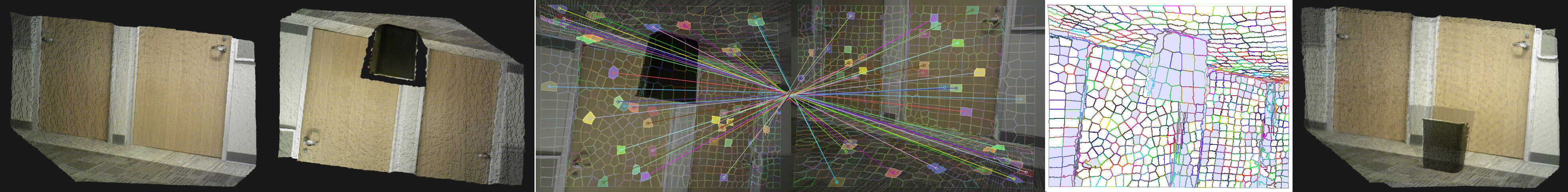}
}\end{subfigure}\hfill
\vspace*{\spaceabovecaption}
\begin{subfigure}[h]{1\linewidth}
{\caption{\label{fig:qual:umdbed} {Results over a scene from \emph{UMD-Hotel} dataset from \cite{xiao2013sun3d}. The views have significant change in perspective, surface resolution scale, and a partial overlap.}\vspace*{\spacebelowcaption} }
\includegraphics[max width=\linewidth, width=\linewidth,keepaspectratio]{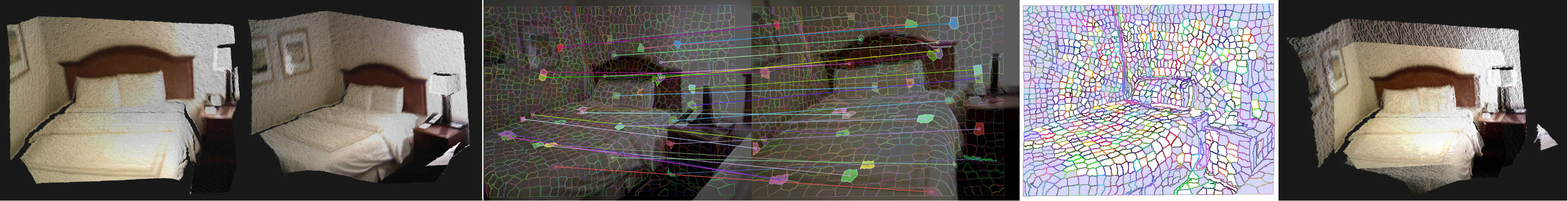}
}\end{subfigure}\hfill
\vspace*{\spaceabovecaption}
\begin{subfigure}[h]{1\linewidth}
{\caption{\label{fig:qual:hvdctable} {Results over a conference room scene from \emph{Harvard-C11} dataset from \cite{xiao2013sun3d}. The views have self-occlusion and partially overlapping geometry.~~~~~~~~~~~~~}\vspace*{\spacebelowcaption} }
\includegraphics[max width=\linewidth, width=\linewidth,keepaspectratio]{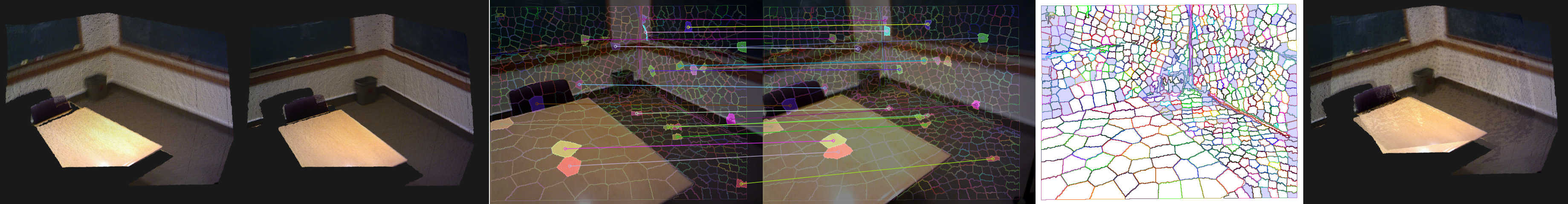}
}\end{subfigure}\hfill
\vspace*{\spaceabovecaption}
\begin{subfigure}[h]{1\linewidth}
{\caption{\label{fig:qual:corridor} {Results over a corridor scene from \emph{Brown-BM1} dataset from \cite{xiao2013sun3d}.~~~~~~~~~~~~~~~~~~~~~~~~~~~~~~~~~~~~~~~~~~~~~~~~~~~~~~~~~~~~~~~~~~~~~~~~~~~~~~~~~~~~~~~~~~~~~~~~~~~~~~~~~~~~~~~~~~~~~~~~~~~~~~}\vspace*{\spacebelowcaption} }
\includegraphics[max width=\linewidth, width=\linewidth,keepaspectratio]{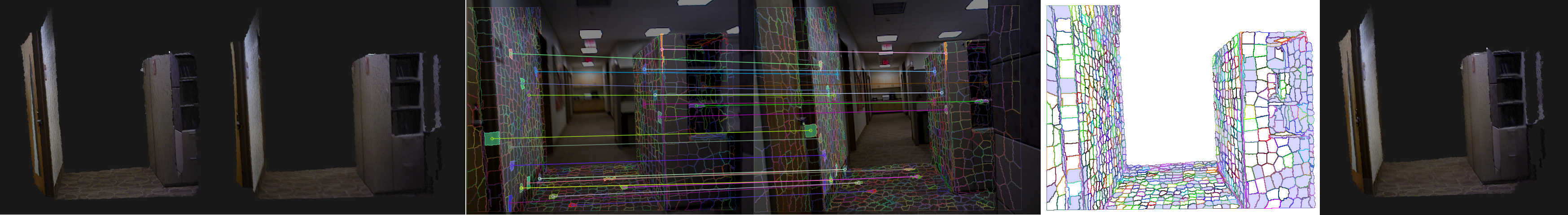}
}\end{subfigure}\hfill
\vspace*{1mm}
\protect\caption{\label{fig:qual} {Example results over varied scenes are shown. These are over different structural settings, and involve varied occlusion, overlap and sensor motion scenarios. Similar presentation and evaluation semantics  as in \emph{Fig.} \emph{\ref{fig:FirstFig}} have been used. Only a sparse sampling of the ascertained associations are indicated in the figures. The achieved associations are nicely localized and densely span the scenes' structures. These include associations over regions which have ambiguous or indiscriminate local geometry (and would prove difficult to associate otherwise).} \vspace*{-1.5mm} }
\end{figure*}

\section{\label{sec:results}Experiments And Results}

We show results on datasets available from \cite{sturm2012RGBDbenchmark,xiao2013sun3d},
as well as ones collected from everyday scenes - these cover a diverse
and challenging range of settings (\emph{Figs.} \emph{\ref{fig:FirstFig}}, \emph{\ref{fig:qual}},
\emph{Table} \emph{\ref{tab:quant}}).

For the experiments in the article, due to the nature of datasets
and to maintain uniformity, full neighborhoods were considered throughout
($k_{S}=|S|,\, k_{S'}=|S'|$). Smaller neighborhoods suffice for settings
with locally anisomorphic content though -- such as ones with clutter.
The superpixel count varied between datasets, and from view to view
(due to regularization) - the average number of superpixels per view
was around 750. KD-tree queries, $C$, were at kept at $75$. The
deviation thresholds, $\theta_{dev}\,\&\, r_{dev}$, were $10^{\circ}\,\&\,.04m$
respectively. 

Exemplar qualitative results are shown in \emph{Figs.} \emph{\ref{fig:FirstFig}}, \emph{\ref{fig:qual}}
- the captions include some major points of note. Nicely localized
associations, densely covering the scenes' structures, were achieved
- this is also indicated by the quality of unrefined reconstructions
resulting from them. As the grey overlays indicate, the occluded and
absent parts always get edited out -- do not get matched. GASP was
able to handle occlusion and partial overlap scenarios, even under
sharp sensor motion. Our experiments indicated the associations obtained
at low gating values to be accurate. Relatively coarser (but qualitatively
correct) associations would arise at higher gating values, with incorrect
associations arising mostly at high $\lambda$ gatings.

We also try to establish the localization precision of GASP's associations,
and its robustness under large sensor motions, through quantitative
evaluations and comparisons for transform estimation accuracies (\emph{Table.}
\emph{\ref{tab:quant}}). Due to space constraints, the evaluation
details and some points of note have been specified within the caption
itself. For automation, roughly chosen high gating values were used
($\lambda\in\{.65,~,.65,~.8\},$ respectively for the three datasets),
and the ensuing associations%
\footnote{A fully dense set of associations was not required for transform estimates.
Instead, associations were only ascertained for a volumetrically downsampled
set of superpixels, uniformly covering the scenes in 3D.%
} were subsequently filtered for 3D consistency through RANSAC - by
simply utilizing the associated superpixels' locations/means as point
correspondences. Note that since we ascertain superpixel level associations,
better transformation estimates could have been obtained by utilizing
richer constraints such as surface patch orientations and overlap
- these were not leveraged in experiments. GASP gave consistently
accurate results, under increasingly large sensor motions and in varied
structural settings. It performed favorably, in comparisons with other
popular approaches.

\begin{table*}[t!]
\vspace*{-2.75mm}
\centering
\bgroup 
\newcommand{\sepwidth}{1pt} 
\newcommand{\shortsepwidth}{1pt} 
\newcommand{\vertsepwidth}{.5pt}
\setlength{\belowrulesep}{0pt} 
\setlength{\aboverulesep}{0pt}
\def\arraystretch{1.15}
\fontsize{7pt}{8pt}\selectfont

\begin{tabular}{cc!{\vrule width \vertsepwidth}cccc!{\vrule width \vertsepwidth}c|c|c|c!{\vrule width \vertsepwidth}ccc}

\specialrule{\sepwidth}{0pt}{1pt}

\multicolumn{2}{c}{{$Datasets~$\cite{sturm2012RGBDbenchmark}}} & \multicolumn{4}{c}{{$Cabinet-SparseStructure$}} & \multicolumn{4}{c}{{$Structure-NoTexture$}} & \multicolumn{3}{c}{{$Household-Clutter$}}\tabularnewline
\hline  
\multicolumn{2}{c}{{$FramesSkipped$}} & {$10$} & {$20$} & {$30$} & {$40$} & {$25$} & {$50$} & {$75$} & {$100$} & {$10$} & {$50$} & {$100$}\tabularnewline 

\specialrule{\sepwidth}{0pt}{1pt}

\multirow{6}{*}{{$\begin{array}{c} Trans_{RMSE}\\ (metres) \end{array}$}} & {$GASP$} & \textbf{\textcolor{red}{0.057}} & \textbf{\textcolor{red}{0.075}} & \textbf{\textcolor{red}{0.070}} & \textbf{\textcolor{red}{0.073}} & \textbf{\textcolor{red}{0.026}} & \textbf{\textcolor{red}{0.037}} & \textbf{\textcolor{red}{0.038}} & \textbf{\textcolor{red}{0.039}} & \textcolor{blue}{0.023} & \textbf{\textcolor{red}{0.028}} & \textbf{\textcolor{red}{0.058}}\tabularnewline \cline{2-13}   & {$SHOT$} & {0.184} & {0.281} & \textcolor{blue}{0.352} & {0.461} & {0.132} & {0.228} & {0.309} & {0.461} & {0.033} & {0.191} & {0.347}\tabularnewline \cline{2-13}   & {$FPFH$} & {0.185} & {0.424 } & {0.393} & \textcolor{blue}{0.435} & {0.164} & {0.202} & {0.275} & {0.406} & {0.045} & {0.260} & {0.606}\tabularnewline

\cline{2-13}
\cmidrule[\shortsepwidth]{2-2}

& {$C-SHOT$} & \textcolor{blue}{0.157}{ } & \textcolor{blue}{0.255} & {0.363} & {0.487} & {0.100} & {0.213} & {0.300} & {0.362} & {0.030} & \textcolor{blue}{0.061} & \textcolor{blue}{0.236}\tabularnewline \cline{2-13}   & {$SIFT$} & {0.201} & {0.285} & {0.405} & {0.468} & \textcolor{blue}{0.046} & \textcolor{blue}{0.059} & \textcolor{green}{0.098} & \textcolor{blue}{0.198} & \textbf{\textcolor{red}{0.014}} & \textcolor{green}{0.029} & \textcolor{green}{0.095}\tabularnewline \cline{2-13}   & {$D-SIFT$} & \textcolor{green}{0.131} & \textcolor{green}{0.178} & \textcolor{green}{0.246} & \textcolor{green}{0.242} & \textcolor{green}{0.030} & \textcolor{green}{0.042} & \textcolor{blue}{0.191} & \textcolor{green}{0.182} & \textcolor{green}{0.019} & \textcolor{black}{0.138} & {0.425}\tabularnewline 

\specialrule{\sepwidth}{0pt}{1pt}

\multirow{6}{*}{{$\begin{array}{c} Rot_{RMSE}\\ (degrees) \end{array}$}} & {$GASP$} & \textbf{\textcolor{red}{1.656}} & \textbf{\textcolor{red}{1.971}} & \textbf{\textcolor{red}{2.737}} & \textbf{\textcolor{red}{2.596}} & \textbf{\textcolor{red}{0.802}} & \textbf{\textcolor{red}{0.998}} & \textbf{\textcolor{red}{1.157}} & \textbf{\textcolor{red}{1.186}} & \textcolor{blue}{1.471} & \textbf{\textcolor{red}{1.077}} & \textbf{\textcolor{red}{2.359}}\tabularnewline \cline{2-13}   & {$SHOT$} & {4.582} & {9.714 } & {9.166} & {13.195} & {4.392} & {7.226} & {8.860} & {9.896} & {1.973} & {6.235} & {14.494}\tabularnewline \cline{2-13}   & {$FPFH$} & {5.375 } & {11.375 } & {10.146} & {11.671} & {5.334} & {6.729} & {7.764} & {14.748} & {2.157} & {8.656} & {22.387}\tabularnewline 

\cline{2-13}
\cmidrule[\shortsepwidth]{2-2}

& {$C-SHOT$} & \textcolor{blue}{4.359} & {7.678 } & {9.231} & {11.666} & {4.177} & {7.212} & {10.571} & {8.592} & {1.511} & \textcolor{blue}{2.775} & \textcolor{blue}{10.927}\tabularnewline \cline{2-13}   & {$SIFT$} & {5.140} & \textcolor{green}{3.764} & \textcolor{blue}{9.090} & \textcolor{blue}{10.287} & \textcolor{blue}{1.436} & \textcolor{blue}{1.908} & \textcolor{green}{3.226} & \textcolor{green}{4.863} & \textbf{\textcolor{red}{0.531}} & \textcolor{green}{1.113} & \textcolor{green}{4.819}\tabularnewline \cline{2-13}   & {$D-SIFT$} & \textcolor{green}{3.175} & \textcolor{blue}{4.578} & \textcolor{green}{5.930} & \textcolor{green}{6.811} & \textcolor{green}{1.315} & \textcolor{green}{1.790} & \textcolor{blue}{6.302} & \textcolor{blue}{6.228} & \textcolor{green}{0.783} & {5.169} & {21.234}\tabularnewline 

\specialrule{\sepwidth}{0pt}{1pt}

\multirow{6}{*}{{$Fail\, Rate$}} & {$GASP$} & \textcolor{red}{0\%} & \textcolor{red}{0\%} & \textcolor{red}{0\%} & \textcolor{red}{0\%} & \textcolor{red}{0\%} & \textcolor{red}{0\%} & \textcolor{red}{0\%} & \textcolor{green}{7.14\%} & \textcolor{red}{0\%} & \textcolor{red}{0\%} & \textcolor{red}{0\%}\tabularnewline \cline{2-13}   & {$SHOT$} & {7.32\%} & {17.07\%} & {16.67\%} & {25.00\%} & \textcolor{red}{0\%} & {3.33\%} & {20.69\%} & {39.29\%} & \textcolor{red}{0\%} & {9.30\%} & {54.29\%}\tabularnewline \cline{2-13}   & {$FPFH$} & \textcolor{green}{1.22\%} & {7.32\%} & \textcolor{blue}{11.11\%} & {27.50\%} & {3.23\%} & {3.33\%} & {17.24\%} & {32.14\%} & \textcolor{red}{0\%} & \textcolor{red}{0\%} & {45.71\%}\tabularnewline
\cline{2-13}
\cmidrule[\shortsepwidth]{2-2}

& {$C-SHOT$} & {8.54\%} & \textcolor{blue}{2.44\%} & {12.96\%} & \textcolor{blue}{15.00\%} & {3.23\%} & {3.33\%} & \textcolor{blue}{3.45\%} & \textcolor{blue}{17.86\%} & \textcolor{red}{0\%} & {4.65\%} & \textcolor{blue}{28.57\%}\tabularnewline \cline{2-13}   & {$SIFT$} & {47.56\%} & {58.54\%} & {61.11\%} & {67.50\%} & {3.23\%} & {3.33\%} & {13.79\%} & {21.43\%} & \textcolor{red}{0\%} & \textcolor{red}{0\%} & \textcolor{red}{0\%}\tabularnewline \cline{2-13}   & {$D-SIFT$} & \textcolor{green}{1.22\%} & \textcolor{red}{0\%} & \textcolor{green}{1.85\%} & \textcolor{green}{2.50\%} & \textcolor{red}{0\%} & \textcolor{red}{0\%} & \textcolor{red}{0\%} & \textcolor{red}{0\%} & \textcolor{red}{0\%} & {6.98\%} & {34.29\%}\tabularnewline 
\specialrule{\sepwidth}{0pt}{1pt}
\end{tabular}
\egroup 
\\[3pt]
\protect\caption{\label{tab:quant}
{We demonstrate the localization accuracy of GASP's superpixel associations by utlizing them for motion estimates, over kinect datasets from \cite{sturm2012RGBDbenchmark} which have ground truths obtained from a motion-capture system. Translation \& Rotation $RMS$ errors and failure rates are shown. For all metrics, lower values are better. Since the datasets had small inter-frame motions, we skipped frames uniformly, starting from regularly spaced initial frames, to simulate significant changes in scene perspectives, sensor baselines and non-overlapping content. The datasets cover different settings - the first two are over varied structural settings with sparse local information, while the last one is over cluttered household/office settings (\emph{Fig.} \ref{fig:qual:house}). 
As can be seen, the transform estimates from GASP associations are accurate. They remain consistent under increasing frame skips, and with minimal failures.\\[0pt] 
\hspace*{1mm} We compare with geometric as well as appearance based 3D feature approaches (ones below short solid lines), in popular use today. $SHOT$ , $FPFH$ (\cite{tombari2010SHOT,rusu2009FPFH}) are point based 3D descriptors based on local geometry; $C-SHOT$ additionally utlizes color information. Dense keypoints ($>2500$) for them were evaluated by volumetrically downsampling the point clouds. Standard $SIFT$ (\cite{lowe2004sift}) was utilized, by back-projecting its keypoints in 3D -  this is prevalent in RGBD based SfM, SLAM approaches (\cite{xiao2013sun3d,henryRGBDMapping2010}). For Dense SIFT, keypoints were taken with a step size of $8$ pixels, at $3$ scales (${1,3,9}$). For all methods, the final feature matches were ascertained by filtering for transformation consistency using RANSAC. We tried dense cloud based direct estimation techniques (\cite{segal2009GICP, Kerl13VO}), but they required short sensor displacements to operate properly. Top values are ordered as ${\color{red}r}{\color{green}g}{\color{blue}b}$. GASP performed best overall. Its associations gave significantly better motion estimates than geometric feature approaches (whose performance deterioted with increasing sensor motion). It also performed better than appearance based approaches, especially under larger sensor motions  and, in settings with little texture.}\vspace*{-1.75mm} }
\end{table*}

\section{Conclusion}

We presented a practical and effective approach towards the problem
of dense superpixel-level data association across views, without requiring
appearance sensing / information. Our approach involved an invariant
representation of relative geometry over superpixel neighborhoods,
and a partial ordering over them - unique to the represented geometry
itself. Robust Damerau-Levenshtein edit distance was leveraged for
matching these ordered representations. The approach exhibited intrinsic
robustness to sensor noise and inexact superpixel decomposition across
views. It was able to perform in settings with wide baselines, occlusions,
partial overlap and significant viewpoint changes. Promising experiment
results were achieved in varied and difficult setups. 

The approach holds further potential in other applications such as
transferring the semantic labels from one view to another, structure
and primitive detection, and co-segmentation. These will be explored
in the future, as well as improvements in the algorithm by leveraging
appearance, and performing experiments in more challenging datasets.

\section*{Acknowledgements}

This work was supported by the Army Research Lab (ARL) MAST CTA project
(329420). Fuxin Li is supported in part by NSF grants IIS 1016772
and IIS 1320348, and ARO-MURI award W911NF-11-1-0046.

\fontsize{7pt}{6pt}\selectfont
{

\bibliographystyle{abbrv}
\bibliography{gasp_bib}

\begin{thebibliography}{10}

\bibitem{arthur2007kmeanspp}
D.~Arthur and S.~Vassilvitskii.
\newblock {K-Means++}: The advantages of careful seeding.
\newblock In {\em ACM-SIAM symposium on Discrete algorithms}, 2007.

\bibitem{bailey2000data}
T.~Bailey, E.~M. Nebot, J.~Rosenblatt, and H.~F. Durrant-Whyte.
\newblock Data association for mobile robot navigation: A graph theoretic
  approach.
\newblock In {\em Robotics and Automation (ICRA)}, 2000.

\bibitem{barnes2010genPatchMatch}
C.~Barnes, E.~Shechtman, D.~B. Goldman, and A.~Finkelstein.
\newblock The generalized patchmatch correspondence algorithm.
\newblock In {\em ECCV}. 2010.

\bibitem{bleyer2010surfaceStereoSoft}
M.~Bleyer, C.~Rother, and P.~Kohli.
\newblock Surface stereo with soft segmentation.
\newblock In {\em Computer Vision and Pattern Recognition (CVPR)}, 2010.

\bibitem{bodis14piecewiseplanarmodelling}
A.~Bodis-Szomoru, H.~Riemenschneider, and L.~V. Gool.
\newblock Fast, approximate piecewise-planar modeling based on sparse
  structure-from-motion and superpixels.
\newblock In {\em Computer Vision and Pattern Recognition}, 2014.

\bibitem{bustos14fastrotationsearch}
A.~J.~P. Bustos, T.-J. Chin, and D.~Suter.
\newblock Fast rotation search with stereographic projections for {3D}
  registration.
\newblock In {\em Computer Vision and Pattern Recognition (CVPR)}, 2014.

\bibitem{caelli2005graphicalmodelsGraphmatch}
T.~Caelli and T.~Caetano.
\newblock Graphical models for graph matching: Approximate models and optimal
  algorithms.
\newblock {\em Pattern Recognition Letters}, 2005.

\bibitem{cagniart2010probabilisticSurfaceTracking}
C.~Cagniart, E.~Boyer, and S.~Ilic.
\newblock Probabilistic deformable surface tracking from multiple videos.
\newblock In {\em Computer Vision - ECCV}. 2010.

\bibitem{cho2012progressiveGraphmatch}
M.~Cho and K.~M. Lee.
\newblock Progressive graph matching: Making a move of graphs via probabilistic
  voting.
\newblock In {\em CVPR}, 2012.

\bibitem{conte2004thirtyyrsGraph}
D.~Conte, P.~Foggia, C.~Sansone, and M.~Vento.
\newblock Thirty years of graph matching in pattern recognition.
\newblock {\em International journal of pattern recognition and artificial
  intelligence}, 2004.

\bibitem{damerau1964technique}
F.~J. Damerau.
\newblock A technique for computer detection and correction of spelling errors.
\newblock {\em Communications of the ACM}, 1964.

\bibitem{dellaert2001mcmcEM}
F.~Dellaert.
\newblock {\em Monte-Carlo EM for data-association and its applications in
  computer vision}.
\newblock PhD thesis, Carnegie Mellon University, 2001.

\bibitem{dou2013planematchrgbd}
M.~Dou, L.~Guan, J.-M. Frahm, and H.~Fuchs.
\newblock Exploring high-level plane primitives for indoor 3d reconstruction
  with a hand-held {RGB-D} camera.
\newblock In {\em Computer Vision - ACCV Workshops}. 2013.

\bibitem{eshet2013DCSH}
Y.~Eshet, S.~Korman, E.~Ofek, and S.~Avidan.
\newblock {DCSH} - matching patches in {RGBD} images.
\newblock In {\em Computer Vision (ICCV)}, 2013.

\bibitem{folkesson2007mspace}
J.~Folkesson, P.~Jensfelt, and H.~I. Christensen.
\newblock The {M-space} feature representation for {SLAM}.
\newblock {\em Robotics, IEEE Transactions on}, 2007.

\bibitem{Gallup10peicwiseRecon}
D.~Gallup, J.~M. Frahm, and M.~Pollefeys.
\newblock Piecewise planar and non-planar stereo for urban scene
  reconstruction.
\newblock In {\em Computer Vision and Pattern Recognition (CVPR)}, 2010.

\bibitem{gelfand2005robustglobalRegist}
N.~Gelfand, N.~J. Mitra, L.~J. Guibas, and H.~Pottmann.
\newblock Robust global registration.
\newblock In {\em Symposium on Geometry Processing}, 2005.

\bibitem{hacohen2011nonrigidDenseCorrespond}
Y.~HaCohen, E.~Shechtman, D.~B. Goldman, and D.~Lischinski.
\newblock Non-rigid dense correspondence with applications for image
  enhancement.
\newblock In {\em ACM Transactions on Graphics (TOG)}. ACM, 2011.

\bibitem{henry2013patch}
P.~Henry, D.~Fox, A.~Bhowmik, and R.~Mongia.
\newblock Patch volumes: Segmentation-based consistent mapping with {RGB-D}
  cameras.
\newblock In {\em 3D Vision (3DV), International Conference on}, 2013.

\bibitem{henryRGBDMapping2010}
P.~Henry, M.~Krainin, E.~Herbst, X.~Ren, and D.~Fox.
\newblock {RGB-D} mapping: Using depth cameras for dense {3D} modeling of
  indoor environments.
\newblock In {\em International Symposium on Experimental Robotics}. 2010.

\bibitem{herbst2013rgbdFlow}
E.~Herbst, X.~Ren, and D.~Fox.
\newblock {RGB-D} flow: Dense {3D} motion estimation using color and depth.
\newblock In {\em ICRA}. IEEE, 2013.

\bibitem{huang2011visual}
A.~S. Huang, A.~Bachrach, P.~Henry, M.~Krainin, D.~Maturana, D.~Fox, and
  N.~Roy.
\newblock Visual odometry and mapping for autonomous flight using an {RGB-D}
  camera.
\newblock In {\em ISRR}, 2011.

\bibitem{Kerl13VO}
C.~Kerl, J.~Sturm, and D.~Cremers.
\newblock Robust odometry estimation for {RGB-D} cameras.
\newblock In {\em Robotics and Automation (ICRA)}. IEEE, 2013.

\bibitem{kowdle2012multiviewCoseg}
A.~Kowdle, S.~N. Sinha, and R.~Szeliski.
\newblock Multiple view object cosegmentation using appearance and stereo cues.
\newblock In {\em ECCV}. 2012.

\bibitem{li20073d}
H.~Li and R.~Hartley.
\newblock The {3D-3D} registration problem revisited.
\newblock In {\em Computer Vision (ICCV)}. IEEE, 2007.

\bibitem{lin2013holisticRgbd}
D.~Lin, S.~Fidler, and R.~Urtasun.
\newblock Holistic scene understanding for {3D} object detection with rgbd
  cameras.
\newblock In {\em ICCV}, 2013.

\bibitem{lowe2004sift}
D.~G. Lowe.
\newblock Distinctive image features from scale-invariant keypoints.
\newblock {\em International journal of computer vision (IJCV)}, 2004.

\bibitem{mivcuvsik2010superpixelStereo}
B.~Mi{\v{c}}u{\v{s}}{\'\i}k and J.~Ko{\v{s}}eck{\'a}.
\newblock Multi-view superpixel stereo in urban environments.
\newblock {\em International Journal of Computer Vision}, 2010.

\bibitem{navarro2001approxStringMatching}
G.~Navarro.
\newblock A guided tour to approximate string matching.
\newblock {\em ACM computing surveys (CSUR)}, 2001.

\bibitem{neira2001JCBB}
J.~Neira and J.~D. Tard{\'o}s.
\newblock Data association in stochastic mapping using the joint compatibility
  test.
\newblock {\em IEEE Robotics and Automation}, 2001.

\bibitem{Newcombe11KFusion}
R.~A. Newcombe, S.~Izadi, O.~Hilliges, D.~Molyneaux, D.~Kim, A.~J. Davison,
  P.~Kohli, J.~Shotton, S.~Hodges, and A.~Fitzgibbon.
\newblock Kinectfusion: Real-time dense surface mapping and tracking.
\newblock In {\em International Symposium on Mixed and Augmented Reality
  (ISMAR)}, 2011.

\bibitem{olson2005singleclustSpectralGraph}
E.~Olson, M.~Walter, S.~J. Teller, and J.~J. Leonard.
\newblock Single-cluster spectral graph partitioning for robotics applications.
\newblock In {\em RSS}, 2005.

\bibitem{olsson2006registration}
C.~Olsson, F.~Kahl, and M.~Oskarsson.
\newblock The registration problem revisited: Optimal solutions from points,
  lines and planes.
\newblock In {\em Computer Vision and Pattern Recognition (CVPR)}, 2006.

\bibitem{pandey2011visbootstrapGICP}
G.~Pandey, J.~R. McBride, S.~Savarese, and R.~M. Eustice.
\newblock Visually bootstrapped generalized {ICP}.
\newblock In {\em ICRA}, 2011.

\bibitem{pathak2010fastPlaneregistration}
K.~Pathak, A.~Birk, N.~Vaskevicius, and J.~Poppinga.
\newblock Fast registration based on noisy planes with unknown correspondences
  for {3-D} mapping.
\newblock {\em Robotics, IEEE Transactions on}, 2010.

\bibitem{ren2012rgbdLabelling}
X.~Ren, L.~Bo, and D.~Fox.
\newblock {RGB-D} scene labeling: Features and algorithms.
\newblock In {\em Computer Vision Pattern Recognition (CVPR)}, 2012.

\bibitem{rusu2009FPFH}
R.~B. Rusu, N.~Blodow, and M.~Beetz.
\newblock Fast point feature histograms for {3D} registration.
\newblock In {\em Robotics and Automation (ICRA)}, 2009.

\bibitem{segal2009GICP}
A.~Segal, D.~Haehnel, and S.~Thrun.
\newblock Generalized-{ICP}.
\newblock In {\em Robotics: Science and Systems (RSS)}, 2009.

\bibitem{sinha14stereoLocalPlaneSweeps}
S.~N. Sinha, D.~Scharstein, and R.~Szeliski.
\newblock Efficient high-resolution stereo matching using local plane sweeps.
\newblock In {\em Computer Vision and Pattern Recognition (CVPR)}, 2014.

\bibitem{stoyanov20123D-NDT}
T.~D. Stoyanov, M.~Magnusson, H.~Andreasson, and A.~Lilienthal.
\newblock Fast and accurate scan registration through minimization of the
  distance between compact {3D NDT} representations.
\newblock {\em The International Journal of Robotics Research (IJRR)}, 2012.

\bibitem{sturm2012RGBDbenchmark}
J.~Sturm, N.~Engelhard, F.~Endres, W.~Burgard, and D.~Cremers.
\newblock A benchmark for the evaluation of {RGB-D SLAM} systems.
\newblock In {\em Intelligent Robots and Systems (IROS)}, 2012.

\bibitem{taguchi2013point}
Y.~Taguchi, Y.-D. Jian, S.~Ramalingam, and C.~Feng.
\newblock Point-plane {SLAM} for hand-held {3D} sensors.
\newblock In {\em Robotics and Automation}, 2013.

\bibitem{thrun1998probabilistic}
S.~Thrun, W.~Burgard, and D.~Fox.
\newblock A probabilistic approach to concurrent mapping and localization for
  mobile robots.
\newblock {\em Autonomous Robots}, 1998.

\bibitem{tombari2010SHOT}
F.~Tombari, S.~Salti, and L.~Di~Stefano.
\newblock Unique signatures of histograms for local surface description.
\newblock In {\em ECCV}. 2010.

\bibitem{tombari2011combined}
F.~Tombari, S.~Salti, and L.~Di~Stefano.
\newblock A combined texture-shape descriptor for enhanced {3D} feature
  matching.
\newblock In {\em Image Processing (ICIP), International Conference on}, 2011.

\bibitem{torresani2013dualdecomp}
L.~Torresani, V.~Kolmogorov, and C.~Rother.
\newblock A dual decomposition approach to feature correspondence.
\newblock {\em Pattern Analysis and Machine Intelligence (PAMI)}, 2013.

\bibitem{trevor2012planarSlam}
A.~J. Trevor, J.~Rogers, and H.~I. Christensen.
\newblock Planar surface {SLAM} with {3D} and {2D} sensors.
\newblock In {\em Robotics and Automation (ICRA)}, 2012.

\bibitem{Wang2013densitymaximGraphMatch}
C.~Wang, L.~Wang, and L.~Liu.
\newblock Improving graph matching via density maximization.
\newblock In {\em Computer Vision (ICCV)}, 2013.

\bibitem{whelan13deformationbasedLC}
T.~Whelan, M.~Kaess, J.~Leonard, and J.~McDonald.
\newblock Deformation-based loop closure for large scale dense {RGB-D SLAM}.
\newblock In {\em IROS}, 2013.

\bibitem{xiao2013sun3d}
J.~Xiao, A.~Owens, and A.~Torralba.
\newblock {SUN3D}: A database of big spaces reconstructed using {SfM} and
  object labels.
\newblock In {\em Computer Vision (ICCV), International Conference on}, 2013.

\bibitem{yang2013go}
J.~Yang, H.~Li, and Y.~Jia.
\newblock {Go-ICP} : Solving {3D} registration efficiently and globally
  optimally.
\newblock In {\em Computer Vision (ICCV)}, 2013.

\end{thebibliography}

}
\end{document}